\documentclass[11pt,english,twocolumn]{article} 
\pdfoutput=1
\usepackage{acronym}
\usepackage{amsmath, amsthm, amssymb, amsfonts, mathtools}
\usepackage{babel}
\usepackage{color}
\usepackage{float}
\usepackage{graphicx}
\usepackage{sidecap}
\usepackage{subfig}
\usepackage[T1]{fontenc}
\usepackage{tablefootnote}
\usepackage{units}
\usepackage{url,lineno}
\usepackage{verbatim}
\usepackage{times}
\usepackage[affil-it]{authblk}
\usepackage[twocolumn,textwidth=19.2cm,columnsep=1.0cm]{geometry}

\graphicspath{{img/}{./}}

\renewcommand{\refeq}[1]{{Eq.~(\ref{#1})}}

\newcommand{\reffig}[1]{{Fig.~\ref{#1}}}
\newcommand{\reftab}[1]{{Tab.~\ref{#1}}}

\newcommand{\refsec}[1]{{Sec.~\ref{#1}}}

\newcommand{\citepd}[1]{\cite{#1}}
\newcommand{\citeauthor}[1]{\cite{#1}}

\author[1]{Emre Neftci \thanks{Electronic address: \texttt{nemre@ucsd.edu}; Corresponding author}}
\author[1,2]{Srinjoy Das}
\author[3]{Bruno Pedroni}
\author[2]{Kenneth Kreutz-Delgado}
\author[1,3]{Gert Cauwenberghs}

\affil[1]{Institute for Neural Computation, UCSD, La Jolla, CA, USA}
\affil[2]{Electrical and Computer Engineering Department, UCSD, La Jolla, CA, USA}
\affil[3]{Department of Bioengineering, UCSD, La Jolla, CA, USA}

\acrodef{AC}[AC]{Arrenhius \& Current}
\acrodef{AER}[AER]{Address Event Representation}
\acrodef{AEX}[AEX]{AER EXtension board}
\acrodef{AMDA}[AMDA]{``AER Motherboard with D/A converters''}
\acrodef{API}[API]{Application Programming Interface}
\acrodef{BM}[BM]{Boltzmann Machine}
\acrodef{CAVIAR}[CAVIAR]{Convolution AER Vision Architecture for Real-Time}
\acrodef{CCN}[CCN]{Cooperative and Competitive Network}
\acrodef{CD}[CD]{Contrastive Divergence}
\acrodef{CMOS}[CMOS]{Complementary Metal--Oxide--Semiconductor}
\acrodef{COTS}[COTS]{Commercial Off-The-Shelf}
\acrodef{CPU}[CPU]{Central Processing Unit}
\acrodef{CV}[CV]{Coefficient of Variation}
\acrodef{CV}[CV]{Coefficient of Variation}
\acrodef{DAC}[DAC]{Digital--to--Analog}
\acrodef{DBN}[DBN]{Deep Belief Network}
\acrodef{DFA}[DFA]{Deterministic Finite Automaton}
\acrodef{DFA}[DFA]{Deterministic Finite Automaton}
\acrodef{divmod3}[DIVMOD3]{divisibility of a number by 3}
\acrodef{DPE}[DPE]{Dynamic Parameter Estimation}
\acrodef{DPI}[DPI]{Differential-Pair Integrator}
\acrodef{DSP}[DSP]{Digital Signal Processor}
\acrodef{DVS}[DVS]{Dynamic Vision Sensor}
\acrodef{EDVAC}[EDVAC]{Electronic Discrete Variable Automatic Computer}
\acrodef{EIF}[EI\&F]{Exponential Integrate \& Fire}
\acrodef{EIN}[EIN]{Excitatory--Inhibitory Network}
\acrodef{EPSC}[EPSC]{Excitatory Post-Synaptic Current}
\acrodef{EPSP}[EPSP]{Excitatory Post--Synaptic Potential}
\acrodef{FPGA}[FPGA]{Field Programmable Gate Array}
\acrodef{FSM}[FSM]{Finite State Machine}
\acrodef{GPU}[GPU]{Graphical Processing Unit}
\acrodef{HAL}[HAL]{Hardware Abstraction Layer}
\acrodef{HH}[H\&H]{Hodgkin \& Huxley}
\acrodef{HMM}[HMM]{Hidden Markov Model}
\acrodef{HW}[HW]{Hardware}
\acrodef{hWTA}[hWTA]{Hard Winner--Take--All}
\acrodef{IF2DWTA}[IF2DWTA]{Integrate \& Fire 2--Dimensional WTA}
\acrodef{IF}[I\&F]{Integrate \& Fire}
\acrodef{IFSLWTA}[IFSLWTA]{Integrate \& Fire Stop Learning WTA}
\acrodef{INCF}[INCF]{International Neuroinformatics Coordinating Facility}
\acrodef{INI}[INI]{Institute of Neuroinformatics}
\acrodef{IO}[IO]{Input-Output}
\acrodef{IPSC}[IPSC]{Inhibitory Post-Synaptic Current}
\acrodef{ISI}[ISI]{Inter--Spike Interval}
\acrodef{JFLAP}[JFLAP]{Java - Formal Languages and Automata Package}
\acrodef{LIF}[LI\&F]{Linear Integrate \& Fire}
\acrodef{LSM}[LSM]{Liquid State Machine}
\acrodef{LTD}[LTD]{Long-Term Depression}
\acrodef{LTI}[LTI]{Linear Time-Invariant}
\acrodef{LTP}[LTP]{Long-Term Potentiation}
\acrodef{LTU}[LTU]{Linear Threshold Unit}
\acrodef{MCMC}{Markov Chain Monte Carlo}
\acrodef{NHML}[NHML]{Neuromorphic Hardware Mark-up Language}
\acrodef{NMDA}[NMDA]{NMDA}
\acrodef{NME}[NE]{Neuromorphic Engineering}
\acrodef{PCB}[PCB]{Printed Circuit Board}
\acrodef{PRC}[PRC]{Phase Response Curve}
\acrodef{PSC}[PSC]{Post-Synaptic Current}
\acrodef{PSP}[PSP]{Post--Synaptic Potential}
\acrodef{RI}[KL]{Kullback-Leibler}
\acrodef{RRAM}[RRAM]{Resistive Random-Access Memory}
\acrodef{RBM}[RBM]{Restricted Boltzmann Machine}
\acrodef{ROC}[ROC]{Receiver Operator Characteristic}
\acrodef{SAC}[SAC]{Selective Attention Chip}
\acrodef{SCD}[SCD]{Spike-Based Contrastive Divergence}
\acrodef{SCX}[SCX]{Silicon CorteX}
\acrodef{STDP}[STDP]{Spike Time Dependent Plasticity}
\acrodef{SW}[SW]{Software}
\acrodef{sWTA}[SWTA]{Soft Winner--Take--All}
\acrodef{VHDL}[VHDL]{VHSIC Hardware Description Language}
\acrodef{VLSI}[VLSI]{Very  Large  Scale  Integration}
\acrodef{WTA}[WTA]{Winner--Take--All}
\acrodef{XML}[XML]{eXtensible Mark-up Language}

\begin{document}

\title{Event-Driven Contrastive Divergence for Spiking Neuromorphic Systems}
\maketitle
\begin{abstract}
\acp{RBM} and Deep Belief Networks have been demonstrated to perform efficiently in a variety of applications, such as dimensionality reduction, feature learning, and classification.
Their implementation on neuromorphic hardware platforms emulating large-scale networks of spiking neurons can have significant advantages from the perspectives of scalability, power dissipation and real-time interfacing with the environment.
However the traditional \ac{RBM} architecture and the commonly used training algorithm known as \ac{CD} are based on discrete updates and exact arithmetics which do not directly map onto a dynamical neural substrate.
Here, we present an event-driven variation of \ac{CD} to train a \ac{RBM} constructed with \ac{IF} neurons, that is constrained by the limitations of existing and near future neuromorphic hardware platforms. 
Our strategy is based on neural sampling, which allows us to synthesize a spiking neural network that samples from a target Boltzmann distribution. 
The recurrent activity of the network replaces the discrete steps of the \ac{CD} algorithm, while \ac{STDP} carries out the weight updates in an online, asynchronous fashion.\\
We demonstrate our approach by training an \ac{RBM} composed of leaky \ac{IF} neurons with \ac{STDP} synapses to learn a generative model of the MNIST hand-written digit dataset, and by testing it in recognition, generation and cue integration tasks.\\
Our results contribute to a machine learning-driven approach for synthesizing networks of spiking neurons capable of carrying out practical, high-level functionality.
\end{abstract}

\section{Introduction}

Machine learning algorithms based on stochastic neural network models such as \acp{RBM} and deep networks are currently the state-of-the-art in several practical tasks \citepd{Hinton_Salakhutdinov06,Bengio09}.
The training of these models requires significant computational resources, and is often carried out using power-hungry hardware such as large clusters \citepd{Le_etal11} or graphics processing units \citepd{Bergstra_etal10}.
Their implementation in dedicated hardware platforms can therefore be very appealing from the perspectives of power dissipation and of scalability.

Neuromorphic \ac{VLSI} systems exploit the physics of the device to emulate very densely the performance of biological neurons in a real-time fashion, while dissipating very low power~\citepd{Mead89,Indiveri_etal11}.
The distributed structure of \acp{RBM} suggests that neuromorphic \ac{VLSI} circuits and systems can become ideal candidates for such a platform. Furthermore, the communication between neuromorphic components is often mediated using asynchronous address-events \citepd{Deiss_etal98} enabling them to be interfaced with event-based sensors \citepd{Liu_Delbruck10,OConnor_etal13,Neftci_etal13c} for embedded applications, and to be implemented in a very scalable fashion \citepd{Joshi_etal10,Silver_etal07,Schemmel_etal10}.

Currently, \acp{RBM} and the algorithms used to train them are designed to operate efficiently on digital processors, using batch, discrete-time, iterative updates based on exact arithmetic calculations.
However, unlike digital processors, neuromorphic systems compute through the continuous-time dynamics of their components, which are typically \acf{IF} neurons \citepd{Indiveri_etal11}, rendering the transfer of such algorithms on such platforms a non-trivial task. 
We propose here a method to construct \acp{RBM} using \ac{IF} neuron models and to train them using an online, event-driven adaptation of the \acf{CD} algorithm.

We take inspiration from computational neuroscience to identify an efficient neural mechanism for sampling from the underlying probability distribution of the \ac{RBM}.
Neuroscientists argue that brains deal with uncertainty in their environments by encoding and combining probabilities optimally \citepd{Doya_etal07}, and that such computations are at the core of cognitive function \citepd{Griffiths_etal10}.
While many mechanistic theories of how the brain might achieve this exist, a recent \emph{neural sampling} theory postulates that the spiking activity of the neurons encodes samples of an underlying probability distribution \citepd{Fiser_etal10}. 
The advantage for a neural substrate in using such a strategy over the alternative one, in which neurons encode probabilities, is that it requires exponentially fewer neurons.
Furthermore, abstract model neurons consistent with the behavior of biological neurons can implement \ac{MCMC} sampling \citepd{Buesing_etal11}, and \acp{RBM} sampled in this way can be efficiently trained using \ac{CD}, with almost no loss in performance \citepd{Pedroni_etal13}.
We identify the conditions under which a dynamical system consisting of \ac{IF} neurons performs neural sampling. These conditions are compatible with neuromorphic implementations of \ac{IF} neurons, suggesting that they can achieve similar performance.
%
The calibration procedure necessary for configuring the parameters of the spiking neural network is based on firing rate measurements, and so is easy to realize in software and in hardware platforms.

In standard \ac{CD}, weight updates are computed on the basis of alternating, feed-forward propagation of activities \citepd{Hinton02}.
In a neuromorphic implementation, this translates to reprogramming the network connections and resetting its state variables at every step of the training. As a consequence, it requires two distinct dynamical systems: one for normal operation (\emph{i.e.} testing), the other for training, which is highly impractical.
To overcome this problem, we train the neural \acp{RBM} using an online adaptation of \ac{CD}. We exploit the recurrent structure of the network to mimic the discrete ``construction'' and ``reconstruction'' steps of \ac{CD} in a spike-driven fashion, and \acf{STDP} to carry out the weight updates. Each sample (spike) of each random variable (neuron) causes synaptic weights to be updated. We show that, over longer periods of time, these microscopic updates behave like a macroscopic \ac{CD} weight update. 
Compared to standard \ac{CD}, no additional programming overhead is required during the training steps, and both testing and training take place in the same dynamical system.

Because \acp{RBM} are generative models, they can act simultaneously as classifiers, content-addressable memories, and carry out probabilistic inference.
We demonstrate these features in a MNIST hand-written digit task \citepd{LeCun_etal98}, using an \ac{RBM} network consisting of $824$ ``visible`` neurons and $500$ ``hidden'' neurons. The spiking neural network was able to learn a generative model capable of recognition performances with accuracies up to $91.9\%$, which is close to the performance obtained using standard \ac{CD} and Gibbs sampling, $93.6\%$.
\section{Materials and Methods}
\subsection{Neural Sampling with Noisy \acs{IF} Neurons}\label{sec:ns}
We describe here the conditions under which a dynamical system composed of \ac{IF} neurons can perform neural sampling.
In has been proven that abstract neuron models consistent with the behavior of biological spiking neurons can perform \ac{MCMC} sampling of a Boltzmann distribution \citepd{Buesing_etal11}. Two conditions are sufficient for this. First, the instantaneous firing rate of the neuron verifies:
\begin{equation} \label{eq:refr_exp_hazard}
    \rho(u(t), t-t') =  
      \begin{dcases*}
        0 & if $t-t'<\tau_{r}$\\
        r(u(t)) & $t-t' \geq \tau_{r}$
      \end{dcases*},
\end{equation}
with $r(u(t))$ proportional to $\exp(u(t))$, where $u(t)$ is the membrane potential and $\tau_{r}$ is an absolute refractory period during which the neuron cannot fire. $\rho(u(t), t-t')$ describes the neuron's instantaneous firing rate as a function of $u(t)$ at time $t$, given that the last spike occurred at $t'$. The average firing rate of this neuron model for stationary $u(t)$ is the sigmoid function:
\begin{equation}\label{eq:refr_exp_hazard_tau}
    \rho(u) = (\tau_r+\exp(-u))^{-1}.
\end{equation}
Second, the membrane potential of neuron $i$ is equal to the linear sum of its inputs: 
\begin{equation}\label{eq:vmem_exact}
    u_i(t)=b_i+\sum_{j=1}^{N} w_{ij} z_j(t), \forall i=1,...,N,
\end{equation}
where $b_i$ is a constant bias, and $z_j(t)$ represents the pre-synaptic spike train produced by neuron $j$ and is set to 1 for a duration $\tau_r$ after the neuron has spiked. The terms $w_{ij} z_j(t)$ are identified with the time course of the \ac{PSP}, \emph{i.e.} the response of the membrane potential to a pre-synaptic spike.
The two conditions above define a neuron model, to which we refer as the ``abstract neuron model''.
The network then samples from a Boltzmann distribution:
\begin{equation}\label{eq:boltzmann_distr}
    \begin{split}
        p(z_1, ...., z_k) = &   \frac{1}{Z} \exp \big( - E(z_1, ..., z_k) \big), \text{with}\\
    E(z_1, ..., z_k) = &  - \frac{1}{2} \sum_{ij} W_{ij} z_i z_j - \sum_i b_i z_i,
    \end{split}
\end{equation}
where $Z$ is the partition function, and $E(z_1, ..., z_k)$ can be interpreted as an energy function \citepd{Haykin99}.

An important fact of the abstract neuron model is that, according to the dynamics of $z_j(t)$, the \acp{PSP} are ``rectangular'' and non-additive.
The implementation of a large number of synapses producing such \acp{PSP} is very difficult to realize in hardware, when compared to first-order linear filters that result in ``alpha''-shaped \acp{PSP} \citepd{Destexhe_etal98,Bartolozzi_Indiveri07}.
This is because, in the latter model, the synaptic dynamics are linear, such that a single hardware synapse can be used to generate the same current that would be generated by an arbitrary number of synapses (see also next section).
As a consequence, we will use alpha-shaped \acp{PSP} instead of rectangular \acp{PSP} in our models. The use of the alpha \ac{PSP} over the rectangular \ac{PSP} is the major source of degradation in sampling performance, as we will discuss in \refsec{sec:kldivergence}.

\paragraph{Stochastic \acs{IF} Neurons.}
A neuron whose instantaneous firing rate is consistent with \refeq{eq:refr_exp_hazard} can perform neural sampling. 
\refeq{eq:refr_exp_hazard} is a generalization of the Poisson process to the case when the firing probability depends on the time of the last spike (\emph{i.e.} it is a renewal process), and so can be verified only if the neuron fires stochastically \citepd{Cox62}. 
Stochasticity in \ac{IF} neurons can be obtained through several mechanisms, such as a noisy reset potential, noisy firing threshold, or noise injection \citepd{Plesser_Gerstner00}.
The first two mechanisms necessitate stochasticity in the neuron's parameters, and therefore may require specialized circuitry. 
But noise injection in the form of background Poisson spike trains requires only synapse circuits, which are present in many neuromorphic \ac{VLSI} implementation of spiking neurons \citepd{Indiveri_etal11,Bartolozzi_Indiveri07}.
Furthermore, Poisson spike trains can be generated self-consistently in balanced excitatory-inhibitory networks \citepd{Vreeswijk_Sompolinsky96}, or using finite-size effects and neural mismatch \citepd{Amit_Brunel97}.

We show that the abstract neuron model in \refeq{eq:refr_exp_hazard} can be realized in a simple dynamical system consisting of leaky \ac{IF} neurons with noisy currents.
The neuron's membrane potential below firing threshold $\theta$ is governed by the following differential equation:
\begin{equation}\label{eq:clif}
    \begin{split}
        C \frac{\mathrm{d}}{\mathrm{d}t} u_i & = - g_L u_i + I_i(t) + \sigma \xi(t),\quad u_i(t)\in(-\infty,\theta),\\
    \end{split}
\end{equation}
where $C$ is a membrane capacitance, $u_i$ is the membrane potential of neuron $i$, $g_L$ is a leak conductance, $\sigma \xi(t)$ is a white noise term of amplitude $\sigma$ (which can for example be generated by background activity), $I_i(t)$ its synaptic current and $\theta$ is the neuron's firing threshold. 
When the membrane potential reaches $\theta$, an action potential is elicited.
After a spike is generated, the membrane potential is clamped to the reset potential $u_{rst}$ for a refractory period $\tau_{r}$.

In the case of the neural \ac{RBM}, the currents $I_i(t)$ depend on the layer the neuron is situated in. 
For a neuron $i$ in layer $v$
\begin{equation}\label{eq:Iv}
    \begin{split}
    I_i(t)  = & I^d_i(t) + I^v_i(t),\\
    \tau_{syn} \frac{\mathrm{d}}{\mathrm{d} t} I^{v}_i  = & -I^{v}_i + \sum_{j=1}^{N_h} q_{h_{ji}} h_j(t) + q_{b_i} b_{v_i}(t),
    \end{split}
\end{equation}
where $I^d_{i}(t)$ is a current representing the data (\emph{i.e.} the external input), $I^v$ is the feedback from the hidden layer activity and the bias, and the $q$'s are the respective synaptic weights.

For a neuron $j$ in layer $h$,
\begin{equation}\label{eq:Ih}
    \begin{split}
    I_j(t)  = & I^h_j(t),\\
    \quad  \tau_{syn} \frac{\mathrm{d}}{\mathrm{d} t} I^{h}_j  = & -I^{h}_j + \sum_{i=1}^{N_v} q_{v_{ij}} v_i(t) + q_{b_j} b_{h_j}(t),
    \end{split}
\end{equation}
where $I^h$ is the feedback from the visible layer, and $b_v(t)$ and $b_h(t)$ are Poisson spike trains implementing the bias. The dynamics of $I^h$ and $I^v$ correspond to a first-order linear filter, so each incoming spike results in \acp{PSP} that rise and decay exponentially (\emph{i.e.} alpha-\ac{PSP}) \citepd{Gerstner_Kistler02}.

Can this neuron neuron verify the conditions required for neural sampling?
The membrane potential is already assumed to be equal to the sum of the \acp{PSP} as required by neural sampling.
So to answer the above question we only need to verify whether \refeq{eq:refr_exp_hazard} holds.
\refeq{eq:clif} is a Langevin equation which can be analyzed using the Fokker-Planck equation \citepd{Gardiner12}.
The solution to this equation provides the neuron's input/output response, \emph{i.e.} its transfer curve (for a review, see \citepd{Renart_etal03}):
\begin{equation}\label{eq:tf}
    \rho(u_0) = \left(\tau_{rf} + \tau_m \sqrt{\pi} \int_{\frac{u_{rst}-u_0}{\sigma_V}}^{\frac{\theta-u_0}{\sigma_V}} \mathrm{d}x \exp(x^2) (1+\mathrm{erf}(x)) \right)^{-1},
\end{equation}
where $\mathrm{erf}$ is the error function (the integral of the normal distribution), $u_0 = \frac{I}{g}$ is the stationary value of the membrane potential when injected by a constant current $I$, $u_{rst}$ is the reset voltage, and $\sigma_V^2(u) = \sigma^2/(g_L C)$. 

\begin{figure}
   \begin{center}
   \includegraphics[width=.22\textwidth]{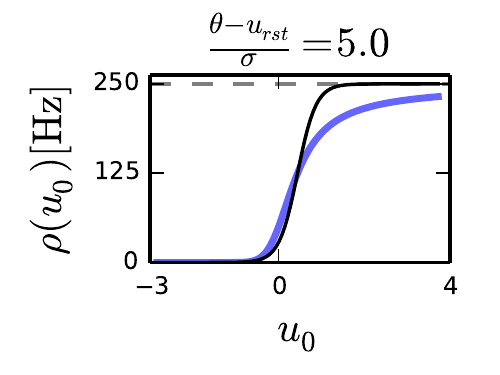} 
   \includegraphics[width=.22\textwidth]{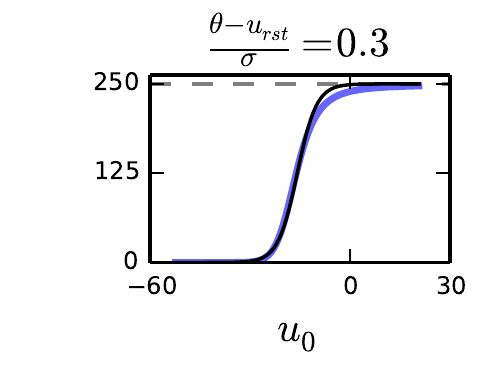}\\ 
   \includegraphics[width=.22\textwidth]{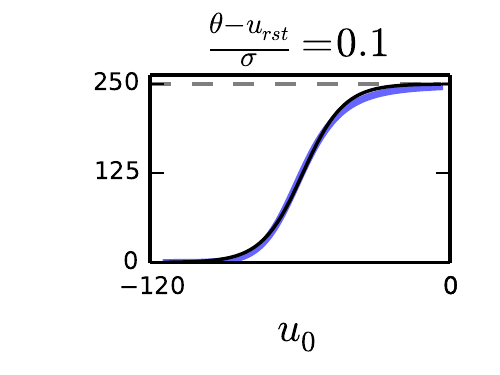} 
   \end{center}
   \caption{\label{fig:tf}
   Transfer curve of a leaky \ac{IF} neuron for three different parameter sets where $u_0 = \frac{I}{g_L}$, and $\frac{1}{\tau_r}=250[Hz]$ (dashed grey).
   In this plot, $\sigma_V$ is varied to produce different ratios  $\frac{\theta-u_{rst}}{\sigma_V}$.
   The three plots above shows that the fit with the sigmoid function (solid black) improves as the ratio decreases.}
\end{figure}

According to \refeq{eq:refr_exp_hazard_tau}, the condition for neural sampling requires that the average firing rate of the neuron to be the sigmoid function.
Although the transfer curve of the noisy \ac{IF} neuron \refeq{eq:tf} is not equal to the sigmoid function, it was previously shown that with an appropriate choice of parameters, the shape of this curve can be very similar to it \citepd{Merolla_etal10}.
We observe that, for a given refractory period $\tau_{r}$, the smaller ratio $\frac{\theta-u_{rst}}{\sigma_V}$ in \refeq{eq:clif}, the better the transfer curve resembles a sigmoid function (\reffig{fig:tf}) .
With a small $\frac{\theta-u_{rst}}{\sigma_V}$, the transfer function of a neuron can be fitted to
\begin{equation}\label{eq:tf_fit}
    \nu(I) = \frac{1}{\tau _{r}} \left(1+\frac{\exp(-I \beta )}{\gamma \tau_{r}}\right)^{-1},
\end{equation}
where $\beta$ and $\gamma$ are the parameters to be fitted.
The choice of the neuron model described in \refeq{eq:clif} is not critical for neural sampling: A relationship that is qualitatively similar to \refeq{eq:tf} holds for neurons with a rigid (reflective) lower boundary \citepd{Fusi_Mattia99} which is common in \ac{VLSI} neurons, and for \ac{IF} neurons with conductance-based synapses \citepd{Petrovici_etal13}.

\begin{figure}
   \begin{center}
   \includegraphics[width=.38\textwidth]{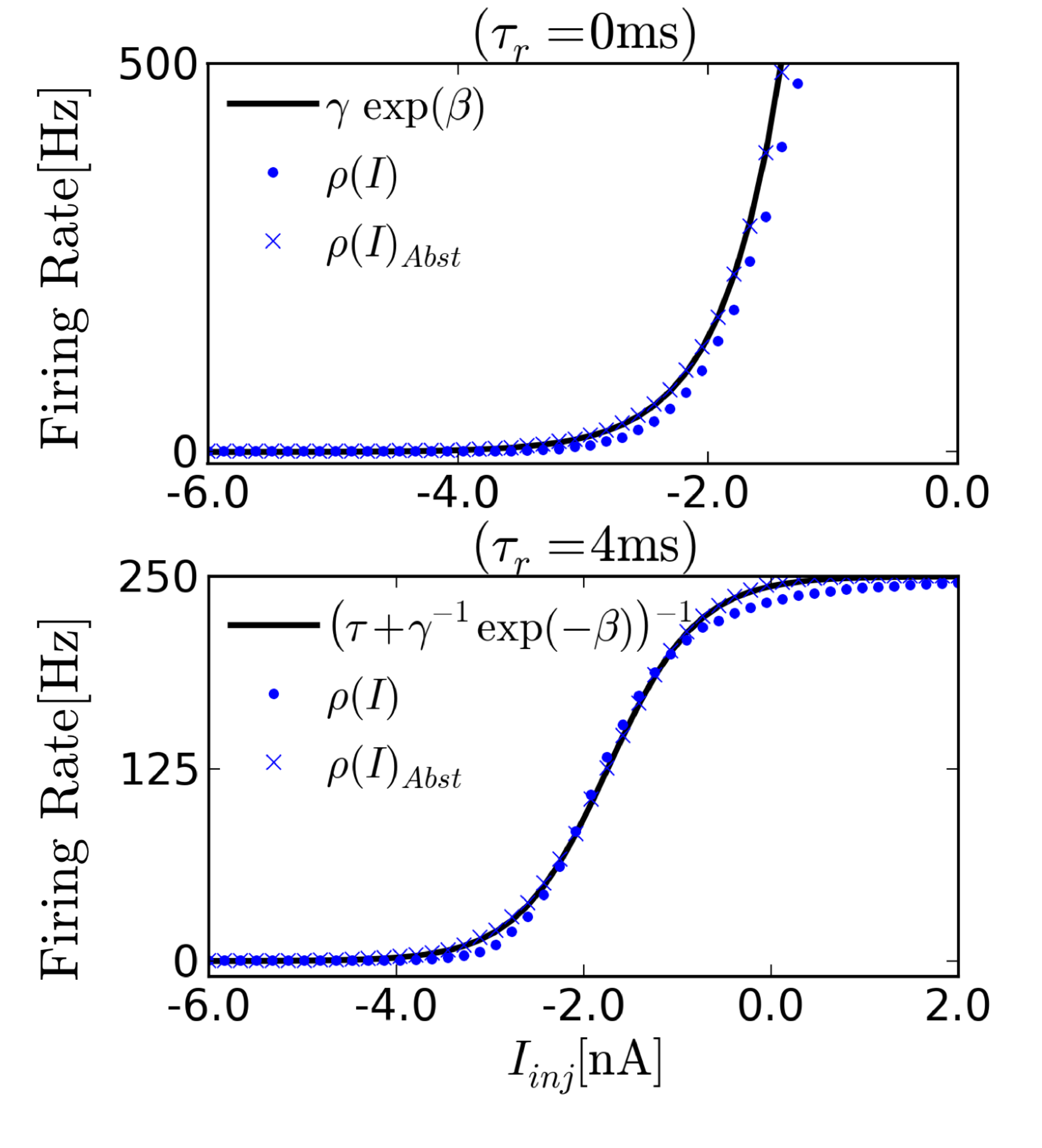} 
   \end{center}
   \caption{\label{fig:calibration}Transfer function of \ac{IF} neurons driven by background white noise (\refeq{eq:clif}). We measure the firing rate of the neuron as a function of a constant current injection to estimate $\rho(u_0)$, where for constant $I_{inj}$, $u_0=I_{inj}/g_{L}$. (Top) The transfer function of noisy \ac{IF} neurons in the absence of refractory period ($\rho(u)= r(u)$, circles). We observe that $\rho$ is approximately exponential over a wide range of inputs, and therefore compatible with neural sampling. Crosses show the transfer curve of neurons implementing the abstract neuron \refeq{eq:refr_exp_hazard}, exactly. (Bottom) With an absolute refractory period the transfer function approximates the sigmoid function. The firing rate saturates at $\unit[250]{Hz}$ due to the refractory period chosen for the neuron. .}
\end{figure}

This result also shows that synaptic weights $q_{v_i}$, $q_{h_j}$, which have the units of charge are related to the \ac{RBM} weights $W_{ij}$ by a factor $\beta^{-1}$.
To relate the neural activity to the Boltzmann distribution, \refeq{eq:boltzmann_distr}, each neuron is associated to a binary random variable which is assumed to take the value $1$ for a duration $\tau_r$ after the neuron has spiked, and zero otherwise, similarly to \cite{Buesing_etal11}. The relation between the random vector and the \ac{IF} neurons' spiking activity is illustrated in \reffig{fig:Example_RBM10}.
\paragraph{Calibration Protocol.}
In order to transfer the parameters from the probability distribution \refeq{eq:boltzmann_distr} to those of the \ac{IF} neurons, the parameters $\gamma$, $\beta$ in \refeq{eq:tf_fit} need to be fitted. 
An estimate of a neuron's transfer function can be obtained by computing its spike rate when injected with different values of constant inputs $I$.
The refractory period $\tau_r$ is the inverse of the maximum firing rate of the neuron, so it can be easily measured by measuring the spike rate for very high input current $I$.
Once $\tau_r$ is known, the parameter estimation can be cast into a simple linear regression problem by fitting $\log(\rho(i)^{-1}-\tau_r)$ with $\beta I + log(\gamma)$.
\reffig{fig:calibration} shows the transfer curve when $\tau_r= \unit[0]{ms}$, which is approximately exponential in agreement with \refeq{eq:refr_exp_hazard}.

The shape of the transfer curse is strongly dependent on the noise amplitude. In the absence of noise, the transfer curve is a sharp threshold function, which softens as the amplitude of the noise is increased (\reffig{fig:tf}). As a result, both parameters $\gamma$ and $\beta$ are dependent on the variance of the input currents from other neurons $I(t)$. Since $\beta q = w$, the effect of the fluctuations on the network is similar to scaling the synaptic weights and the biases which can be problematic.
However, by selecting a large enough noise amplitude $\sigma$ and a slow enough input synapse time constant, the fluctuations due to the background input are much larger than the fluctuations due to the inputs. 
In this case, $\beta$ and $\gamma$ remain approximately constant during the sampling.

Neural mismatch can cause $\beta$ and $\gamma$ to differ from neuron to neuron. From \refeq{eq:tf_fit} and the linearity of the postsynaptic currents $I(t)$ in the weights, it is clear that this type of mismatch can be compensated by scaling the synaptic weights and biases accordingly.
The calibration of the parameters $\gamma$ and $\beta$ quantitatively relate the spiking neural network's parameters to the \ac{RBM}. 
In practice, this calibration step is only necessary for mapping pre-trained parameters of the \ac{RBM} onto the spiking neural network.

Although we estimated the parameters of software simulated \ac{IF} neurons, parameter estimation based on firing rate measurements were shown to be an accurate and reliable method for \ac{VLSI} \ac{IF} neurons as well \citepd{Neftci_etal12}.

\begin{table*}
\begin{center}
    \begin{tabular}{ l | l | c | c }
    \hline
    $\nu_{bias}$  & Mean firing rate of bias Poisson spike train                  & all figures                  & $\unit[1000]{Hz}$\\ 
    $\sigma$      & Noise amplitude                                               & all figures, except \reffig{fig:tf} & $\unit[3\cdot 10^{-11}]{A}$ \\
                  &                                                               & \reffig{fig:tf} (left)       & $\unit[2\cdot 10^{-11}]{A}$\\
                  &                                                               & \reffig{fig:tf} (right)     & $\unit[3\cdot 10^{-10}]{A}$\\
                  &                                                               & \reffig{fig:tf} (bottom)      & $\unit[1\cdot 10^{-9 }]{A}$ \\
    $\beta$       & Exponential factor (fit)                                      & all figures                  & $\unit[2.044\cdot 10^{9}]{A^{-1}}$\\
    $\gamma$      & Baseline firing rate (fit)                                    & all figures                  & $\unit[8808]{Hz}$\\
    $\tau_{r}$    & Refractory period                                             & all figures                  & $\unit[4]{ms}$\\
    $\tau_{syn}$  & Time constant of recurrent, and bias synapses.                & all figures                  & $\unit[4]{ms}$\\
    $\tau_{br}$ & ``Burn-in'' time of the neural sampling                         & all figures                  & $\unit[10]{ms}$\\
    $g_{L}$       & Leak conductance                                              & all figures                  & $\unit[1]{nS}$ \\
    $u_{rst}$     & Reset Potential                                               & all figures                  & $\unit[0]{V}$ \\
    $C$           & Membrane capacitance                                          & all figures                  & $\unit[10^{-12}]{F}$ \\
    $\theta$      & Firing threshold                                              & all figures                  & $\unit[100]{mV}$ \\
    $W$           & \acs{RBM} weight matrix ($\in \mathbb{R}^{N_v \times N_h}$)   & \reffig{fig:kldivergence}    & $N(-.75, 1.5)$ \\ 
    $b_v, b_h$    & \acs{RBM} bias for layer $v$ and $h$                          & \reffig{fig:kldivergence}    & $N(-1.5, .5)$ \\ 
    $N_v, N_h$    & Number of visible and hidden units                            & \reffig{fig:kldivergence}    & $5, 5$ \\ 
                  & in the \acs{RBM}                                              & \reffig{fig:trained_rbm},\ref{tab:performance}     & $824, 500$ \\ 
                  &                                                               & \reffig{fig:classification_reconstruction_inference}  & $834, 500$ \\ 
    $N_c$         & Number of class label units                                   & \reffig{fig:trained_rbm},\ref{tab:performance},\ref{fig:classification_reconstruction_inference}     & $40$ \\ 
    $2T$          & Epoch duration                                                & \reffig{fig:kldivergence}, \ref{fig:trained_rbm},\ref{tab:performance} & $\unit[100]{ms}$\\
                  &                                                               & \reffig{fig:classification_reconstruction_inference} & $\unit[300]{ms}$\\
    $T_{sim}$     & Simulation time                                               & \reffig{fig:calibration}& $\unit[5]{s}$ \\ 
                  &                                                               & \reffig{fig:kldivergence}    & $\unit[1000]{s}$ \\ 
                  &                                                               & \reffig{fig:trained_rbm}    & $\unit[.2]{s}$ \\ 
                  &                                                               & \reffig{fig:classification_reconstruction_inference}    & $\unit[.85]{s}$ \\ 
                  &                                                               & \reffig{tab:performance} (testing)    & $\unit[1.0]{s}$ \\ 
                  &                                                               & \reffig{tab:performance} (learning)   & $\unit[2000]{s}$ \\ 
    $\tau_{STDP}$ & Learning time window                                           &  \reffig{fig:trained_rbm}   & $\unit[4]{ms}$ \\ 
    $\eta$        & Learning rate                                                  & standard \ac{CD}   & $\unit[.1\cdot 10^{-2}]{}$ \\ 
                  &                                                                & event-driven \ac{CD} & $\unit[3.2\cdot 10^{-2}]{}$ \\ 
  \end{tabular}
\end{center}
\caption{\label{tab:parameters}}
\end{table*}
\subsection{Validation of Neural Sampling using \ac{IF} neurons}\label{sec:kldivergence}
\begin{figure}
   \begin{center}
   \includegraphics[width=.5\textwidth]{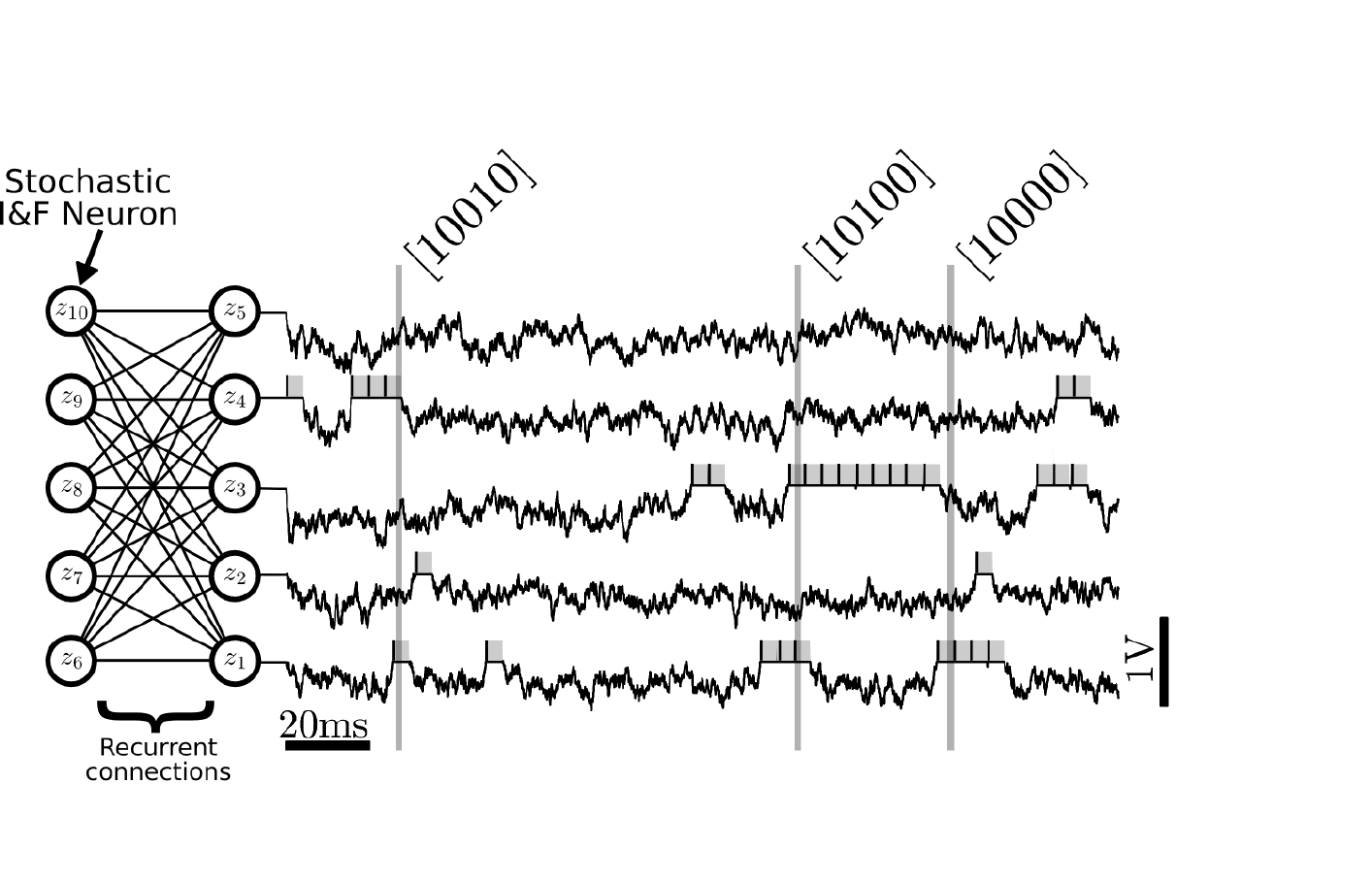} 
   \caption{\label{fig:Example_RBM10} Neural Sampling in an \ac{RBM} consisting of 10 stochastic \ac{IF} neurons, with 5 neurons in each layer. Each neuron is associated to a binary random variable which take values 1 during a refractory period $\tau_r$ after the neuron has spiked (gray shadings). The variables are sampled at $\unit[1]{kHz}$ to produce binary vectors that correspond to samples of the joint distribution $\mathbf{p(\mathbf{z})}$. In this figure, only the membrane potential and the samples produced by the first $5$ neurons are shown. The vectors inside the brackets are example samples of the marginalized distribution $p(z_1,z_2,z_3,z_4,z_5)$ produced at the time indicated by the vertical lines. In the \ac{RBM}, there are no recurrent connections within a layer.}
   \end{center}
\end{figure}

The \ac{IF} neuron verifies \refeq{eq:refr_exp_hazard} only approximately, and the \ac{PSP} model is different than the one of \refeq{eq:vmem_exact}.
Therefore, the following two important questions naturally arise: how accurately does the \ac{IF} neuron-based sampler outlined above sample from a target Boltzmann distribution? How well does it perform in comparison to an exact sampler, such as the Gibbs sampler?
To answer these questions we sample from several neural \ac{RBM} consisting of 5 visible and 5 hidden units for randomly drawn weight and bias parameters. At these small dimensions, the probabilities associated to all possible values of the random vector $\mathbf{z}$ can be computed exactly. These probabilities are then compared to those obtained through the histogram constructed with the sampled events.
To construct this histogram, each spike was extended to form a box of length $\tau_r$ (as illustrated in \reffig{fig:Example_RBM10}), the spiking activity was sampled at \unit[1]{kHz}, and the occurrences of all the possible $2^{10}$ states of the random vector $\mathbf{z}$ were counted.
We added 1 to the number of occurrences of each state to avoid zero probabilities.

A common measure of similarity between two distributions is the \ac{RI} divergence:
\[
    D(p||q)  =  \sum_i p_i \log \frac{p_i}{q_i}.
\]
If the distributions $p$ and $q$ are identical then $D(p||q)=0$, otherwise $D(p||q)>0$.

The average \ac{RI} divergence for 48 randomly drawn distributions after $\unit[1000]{s}$ of sampling time was $\unit[0.058]{}$.
This result is not significantly different if the abstract neuron model \refeq{eq:refr_exp_hazard} with alpha \acp{PSP} is used, and in both cases the \ac{RI} divergence did not tend to zero as the number of samples increased.
The only difference in the latter neuron model compared to the abstract neuron model of \cite{Buesing_etal11}, which tends to zero when sampling time tends to infinity, is the \ac{PSP} model. 
This indicates that the discrepancy is largely due to the use of alpha-\acp{PSP}, rather than the approximation of \refeq{eq:refr_exp_hazard} with \ac{IF} neurons.

The standard sampling procedure used in \acp{RBM} is Gibbs Sampling: the neurons in the visible layer are sampled simultaneously given the activities of the hidden neurons, then the hidden neurons are sampled given the activities of the visible neurons. This procedure is repeated a large number of times. 
For comparison with the neural sampler, the duration of one Gibbs sampling iteration is identified with one refractory period $\tau_r = \unit[4]{ms}$.
At this scale, we observe that the speed of convergence of the neural sampler is similar to that of the Gibbs sampler up to $10^4\unit{ms}$, after which the neural sampler plateaus above the $D(p||q)=10^{-2}$ line.
Despite the approximations in the neuron model and the synapse model, these results show that in \acp{RBM} of this size, the neural sampler consisting of \ac{IF} neurons sample from a distribution that has the same \ac{RI} divergence as the distribution obtained after $10^4$ iterations of Gibbs sampling, which is more than the typical number of iterations used for MNIST hand-written digit tasks in the literature  \citepd{Hinton_etal06}.

\begin{figure*}
   \begin{center}
   \includegraphics[width=.75\textwidth]{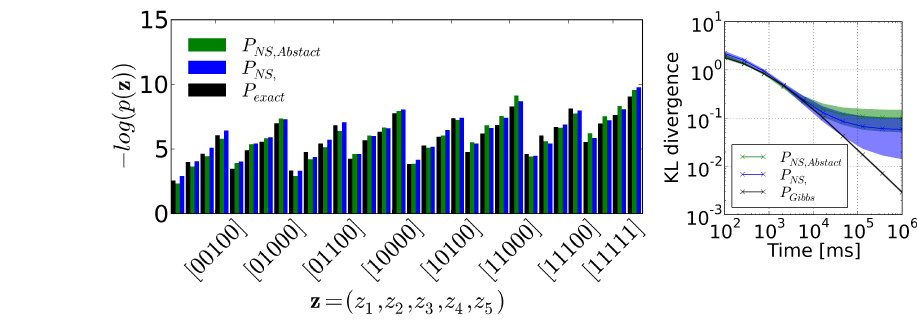} 
    \caption{\label{fig:kldivergence}(Left) Example probability distribution obtained by neural sampling of the \ac{RBM} of \reffig{fig:Example_RBM10}. The bars are marginal probabilities computed by counting the events $[00000],[00001],\cdots,[11110],[11111]$, respectively. $P_{NS}$ is the distribution obtained by neural sampling and $P$ is the exact probability distribution computed with \refeq{eq:boltzmann_distr}.  
    (Right) The degree to which the sampled distribution resembles the target distribution is quantified by the \ac{RI} divergence measured across $\unit[48]{}$ different distributions, and the shadings correspond to its standard deviation. This plot also shows the \ac{RI} divergence of the target distribution sampled by Gibbs Sampling ($P_{Gibbs}$), which is the common choice for \acp{RBM}. For comparison with the neural sampler, we identified the duration of one Gibbs sampling iteration with one refractory period $\tau_r = \unit[4]{ms}$. The plot shows that up to $10^4\unit{ms}$, the two methods are comparable. After this, the \ac{RI} divergence of the neural sampler tends to a plateau due to the fact that neural sampling with our \ac{IF} neural network is approximate.
    In both figures, $P_{NS,Abstract}$ refers to the marginal probability distribution obtained by using the abstract neuron model \refeq{eq:refr_exp_hazard}. In this case, the \ac{RI} divergence is not significantly different from the one obtained with the \ac{IF} neuron model-based sampler.
   }
   \end{center}
\end{figure*}

\subsection{Neural Architecture for Learning a Model of MNIST Hand-Written Digits}\label{sec:nnarch}
We test the performance of the neural \ac{RBM} in a digit recognition task. We use the MNIST database, whose data samples consist of centered, gray-scale, $28\times 28$-pixel images of hand-written digits 0 to 9 \citepd{LeCun_etal98}. 
\begin{figure}
   \begin{center}
   \includegraphics[width=0.45\textwidth]{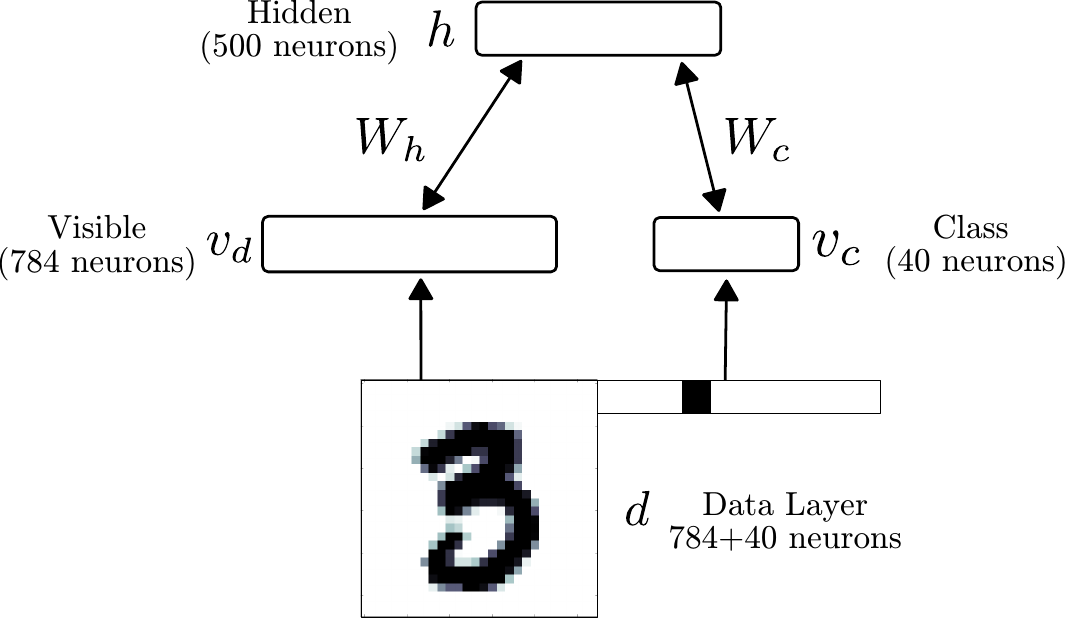}
   \end{center}
   \caption{\label{fig:neural_rbm} The \ac{RBM} network consists of a visible and a hidden layer.  
   The visible layer is partitioned into $784$ sensory neurons ($\mathbf{v_d}$) and $40$ class label neurons ($\mathbf{v_c}$) for supervised learning. During data presentation, the activities in the visible layer are driven by a data layer $\mathbf{d}$, consisting of a digit and its label ($1$ neuron per label). In the \ac{RBM}, the weight matrix between the visible layer and the hidden layer is symmetric.}
\end{figure}                       
The neural \ac{RBM}'s network architecture consisted of 2 layers, as illustrated in \reffig{fig:neural_rbm}. 
The visible layer was partitioned into $784$ sensory neurons ($\mathbf{v_d}$) and $40$ class label neurons ($\mathbf{v_c}$) for supervised learning. 
The pixel values of the digits were discretized to 2 values, with low intensity pixel values ($p<=.5$) mapped to $10^{-5}$ and high intensity values ($p>.5$) mapped to $0.98$.
A neuron $i$ in $\mathbf{d}$ stimulated each neuron $i$ in layer $\mathbf{v}$, with synaptic currents $f_i$ such that $P(v_i = 1) = \nu(f_i) \tau_{r} = p_i$, where $0\le p_i \le 1$ is the value of pixel $i$. The value $f_i$ is calculated by inverting the transfer function of the neuron: $f_i = \nu^{-1}(s) = \log \left(\frac{s}{\gamma-s \gamma \tau_{r}}\right)\beta^{-1}$.
Using this \ac{RBM}, classification is performed by choosing the most likely label given the input, under the learned model. 
This equals to choosing the population of class neurons associated to the same label that has the highest population firing rate.

To reconstruct a digit from a class label, the class neurons belonging to a given digit are clamped to a high firing rate.
For testing the discrimination performance of an energy-based model such as the \ac{RBM}, it is common to compute the free-energy $F(\mathbf{v_c})$ of the class units \citepd{Haykin99}, defined as:
\begin{equation}
    \exp(-F(\mathbf{v_c})) = \sum_{\mathbf{v_d},\mathbf{h}} \exp(-E(\mathbf{v_d},\mathbf{v_c},\mathbf{h})),
\end{equation}
and selecting $\mathbf{v_c}$ such that the free-energy is minimized.
The spiking neural network is simulated using the BRIAN simulator \citepd{Goodman_Brette08}.
All the parameters used in the simulations are provided in \reftab{tab:parameters}.

\section{Results}
\subsection{Event-Driven Contrastive Divergence}
\begin{figure*}
   \begin{center}
   \includegraphics[width=0.7\textwidth]{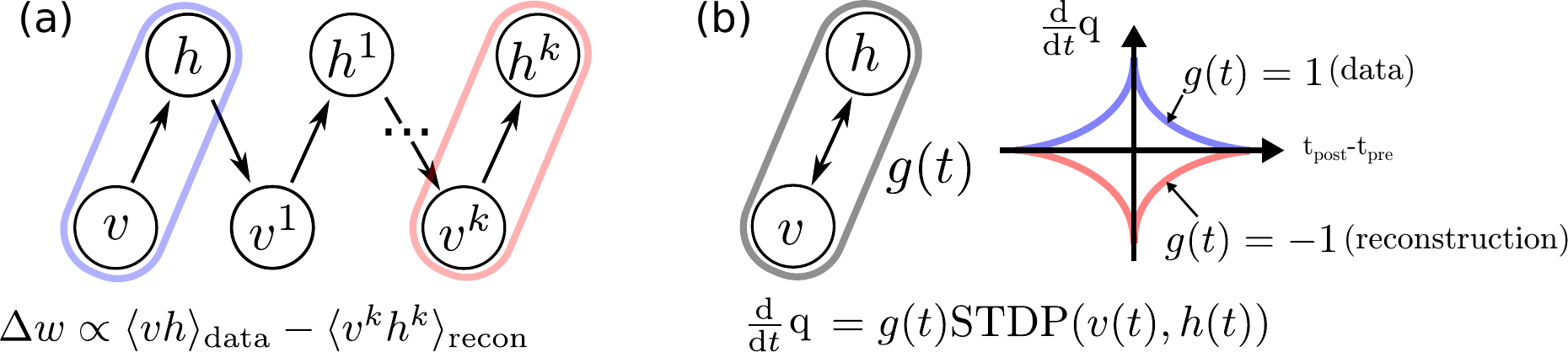}
   \caption{\label{fig:CD_comparison}  The standard \acf{CD}$_k$ procedure, compared to event-driven \ac{CD}. (a) In standard \ac{CD}, learning proceeds iteratively by sampling in ``construction'' and ``reconstruction'' phases~\protect\citepd{Hinton02}, which is impractical in a continuous-time dynamical system. (b) We propose a spiking neural sampling architecture that folds these updates on a continuous time dimension through the recurrent activity of the network. 
   The synaptic weight update follows a \ac{STDP} rule modulated by a zero mean signal $g(t)$. This signal switches the behavior of the synapse from \acf{LTP} to \acf{LTD}, and partitions the training into two phases analogous to those of the original \ac{CD} rule. The spikes cause microscopic weight modifications, which on average behave as the macroscopic \ac{CD} weight update. For this reason, the learning rule is referred to as event-driven \ac{CD}.}
   \end{center}
\end{figure*}
A \acf{RBM} is a stochastic neural network consisting of two symmetrically interconnected layers composed of neuron-like units - a set of visible units $v$ and a set of hidden units $h$, but has no connections within a layer.

The training of \acp{RBM} commonly proceeds in two phases. At first the states of the visible units are clamped to a given vector from the training set, then the states of the hidden units are sampled. In a second ``reconstruction'' phase, the network is allowed to run freely. Using the statistics collected during sampling, the weights are updated in a way that they maximize the likelihood of the data \citepd{Hinton02}.
Collecting equilibrium statistics over the data distribution in the reconstruction phase is often computationally prohibitive.
The \ac{CD} algorithm has been proposed to mitigate this \citepd{Hinton02,Hinton_Salakhutdinov06}: the reconstruction of the visible units activity is achieved by sampling them conditioned on the values of the hidden units (\reffig{fig:CD_comparison}). This procedure can be repeated $k$ times (the rule is then called \ac{CD}$_k$), but relatively good convergence is obtained for the equilibrium distribution even for one iteration. The \ac{CD} learning rule is summarized as follows:
\begin{equation}\label{eq:cd_rule}
    \Delta w_{ij} = \epsilon( \langle v_i h_j \rangle_{\text{data}} - \langle v_i h_j \rangle_{\text{recon}}),
\end{equation}
where $v_i$ and $h_j$ are the activities in the visible and hidden layers, respectively.
This rule can be interpreted as a difference of Hebbian and anti-Hebbian learning rules between the visible and hidden neurons sampled in the data and reconstruction phases. In practice, when the data set is very large, weight updates are calculated using a subset of data samples, or ``mini-batches''. The above rule can then be interpreted as a stochastic gradient descent \citepd{Robbins_Monro51}. Although the convergence properties of the \ac{CD} rule are the subject of continuing investigation, extensive software simulations show that the rule often converges to very good solutions~\citepd{Hinton02}.

The main result of this paper is an online variation of the \ac{CD} rule for implementation in neuromorphic hardware. By virtue of neural sampling the spikes generated from the visible and hidden units can be used to compute the statistics of the probability distributions online (further details on neural sampling in the Materials and Methods \refsec{sec:ns}). Therefore a possible neural mechanism for implementing \ac{CD} is to use synapses whose weights are governed by synaptic plasticity.
Because the spikes cause the weight to update in an online, and asynchronous fashion, we refer to this rule as \emph{event-driven} \ac{CD}.

The weight update in event-driven \ac{CD} is a modulated, pair-based STDP rule:
\begin{equation}\label{eq:ecd_rule}
    \frac{\mathrm{d}}{\mathrm{d}t} q_{ij} = g(t)\,\mathrm{STDP}_{ij}(v_i(t),h_j(t))
\end{equation}
where $g(t) \in \mathbb{R}$ is a zero-mean global gating signal controlling the data vs. reconstruction phase, $q_{ij}$ is the weight of the synapse and $v_i(t)$ and $h_j(t)$ refer to the spike trains of neurons $v_i$ and $h_j$, respectively, which are represented by a sum of Dirac delta pulses centered on the respective spike times:
$
    v_i(t) = \sum_{k\in Sp_i} \delta(t-t_k), \quad h_j(t) = \sum_{k\in Sp_j} \delta(t-t_k)
$
where $Sp_i$ and $Sp_j$ are the set of the spike times of the visible neuron $i$ and hidden neuron $j$, respectively and $\delta(t)=1$ if $t=0$ and $0$ otherwise.

As opposed to the standard \ac{CD} rule, weights are updated after every occurrence of a pre-synaptic and post-synaptic event. 
While this online approach slightly differentiates it from standard \ac{CD}, it is integral to a spiking neuromorphic framework where the data samples and weight updates cannot be stored. 
The weight update is governed by a symmetric \ac{STDP} rule with a symmetric temporal window $K(t)= K(-t), \forall t$: 
\begin{equation}
    \begin{split}
        \mathrm{STDP}_{ij}(v_i(t),h_j(t)) = & v_i(t) A_{h_j}(t) + h_j(t) A_{v_i}(t),\\
A_{h_j}(t)     = & A \int^t_{-\infty} \mathrm{d}s K(s-t) h_j(s),\\
A_{v_i}(t)    = &   A \int^t_{-\infty} \mathrm{d}s K(s-t) v_i(s),\\
    \end{split}
\end{equation} 
with $A>0$ defining the magnitude of the weight updates.
In our implementation, updates are additive and weights can change polarity.
%
\paragraph{Pairwise \acs{STDP} with a global modulatory signal approximates \acs{CD}.}
The modulatory signal $g(t)$ switches the behavior of the synapse from \ac{LTP} to \ac{LTD} (\emph{i.e.} Hebbian to Anti-Hebbian). The temporal average of $g(t)$ must vanish to balance \ac{LTP} and \ac{LTD}, and must vary on much slower time scales than the typical times scale of the network dynamics, denoted $\tau_{br}$, so that the network samples from its stationary distribution when the weights are updated. The time constant $\tau_{br}$ corresponds to a ``burn-in'' time of MCMC sampling and depends on the overall network dynamics and cannot be computed in the general case. However, it is reasonable to assume $\tau_{br}$ to be in the order of a few refractory periods of the neurons \citepd{Buesing_etal11}.
In this work, we used the following modulation function $g(t)$:
\begin{equation}\label{eq:canonical_g}
    g(t) =  
      \begin{dcases*}
          1 & \text{if  $mod(t,2T)\in (\tau_{br}, T)$}\\
          -1 & \text{if  $mod(t,2T) \in (T+\tau_{br}, 2T)$}\\
          0 & \text{otherwise}
      \end{dcases*},
\end{equation}
where $mod$ is the modulo function and $T$ is a time interval.
The data is presented during the time intervals $(2iT,(2i+1)T)$, where $i$ is a positive integer. 
With the $g(t)$ defined above, no weight update is undertaken during a fixed period of time $\tau_{br}$. 
This allows us to neglect the transients after the stimulus is turned on and off (respectively in the beginning of the data and reconstruction phases).
In this case and under further assumptions discussed below, the event-driven \ac{CD} rule can be directly compared with standard \ac{CD} as we now demonstrate.
The average weight update during $(0,2T)$ is:
\begin{equation}
    \begin{split}
        \langle \frac{\mathrm{d}}{\mathrm{d}t} q_{ij} \rangle_{(0,2T)}  = & C_{ij} + R_{ij},\\
C_{ij}  = & \frac{T-\tau_{br}}{2T}(\langle v_i(t) A_{h_j}(t) \rangle_{t_d} + \langle h_j(t) A_{v_i}(t) \rangle_{t_d})\\
R_{ij}  = & - \frac{T-\tau_{br}}{2T}(\langle v_i(t) A_{h_j}(t) \rangle_{t_r} + \langle h_j(t) A_{v_i}(t) \rangle_{t_r}),\\
    \end{split}
\end{equation}
where $t_d = (\tau_{br}, T)$ and $t_r=(T+\tau_{br}, 2T)$ denote the intervals during the positive and negative phases of $g(t)$, and $\langle \cdot \rangle_{(a,b)} = \frac{1}{b-a}\int_{a}^{b} \mathrm{d} t \cdot$. 

We write the first average in $C_{ij}$ as follows:
\begin{equation}\label{eq:cijfirst}
    \begin{split}
        \langle v_i(t) A_{h_j}(t) \rangle_{t_d} 
         = A \frac{1}{T-\tau_{br}} \int_{\tau_{br}}^T \mathrm{d}t \int^t_{-\infty} \mathrm{d}s K(s-t) v_i(t) h_j(s),\\
         = A \frac{1}{T-\tau_{br}} \int_{\tau_{br}}^T \mathrm{d}t \int^0_{-\infty} \mathrm{d}\Delta K(\Delta) \langle v_i(t) h_j(t+\Delta),\\
         = A \int^0_{-\infty}   \mathrm{d}\Delta K(\Delta) \langle v_i(t) h_j(t+\Delta) \rangle_{t_d}.\\
    \end{split}
\end{equation}
If the spike times are uncorrelated the temporal averages become a product of the average firing rates of a pair of visible and hidden neurons \citepd{Gerstner_Kistler02}: 
\[
    \langle v_i(t)h_j(t+\Delta) \rangle_{t_d} = \langle v_i(t) \rangle_{t_d} \langle h_j(t+\Delta) \rangle_{t_d} \eqqcolon \bar{v}_i^+ \bar{h}_j^+.
\]
If we choose a temporal window that is much smaller than $T$, and since the network activity is assumed to be stationary in the interval $(\tau_{br},T)$, we can write (up to a negligible error~\citepd{Kempter_etal01})

\begin{equation}\label{eq:point_symm}
        \langle v_i(t) A_{h_j}(t) \rangle_{t_d} = A \bar{v}_i^+ \bar{h}_j^+ \int^0_{-\infty} \mathrm{d}\Delta K(\Delta).
\end{equation}

In the uncorrelated case, the second term in $C_{ij}$ contributes the same amount, leading to:
\[ 
    C_{ij} = \eta \bar{v}_i^+ \bar{h}_j^+. 
\]
with $\eta \coloneqq 2A \frac{T-\tau_{br}}{2T} \int^{0}_{-\infty} \mathrm{d}{\Delta} K(\Delta)$.
Similar arguments apply to the averages in the time interval $t_r$: 
\[
    R_{ij} = 2 A \int^0_{-\infty} \mathrm{d}\Delta K(\Delta) \langle v_i(t) h_j(t+\Delta) \rangle_{t_r} = \eta \bar{v}_i^- \bar{h}_j^-. 
\]
with $\bar{v}_i^- \bar{h}_j^- \coloneqq \langle v_i(t) \rangle_{t_r} \langle h_j(t+\Delta) \rangle_{t_r}$. 
The average update in $(0,2T)$ then becomes:
\begin{equation}\label{eq:ecd_update}
\langle \frac{\mathrm{d}}{\mathrm{d}t} q_{ij} \rangle_{(0,2T)} = 
\eta \left( \bar{v}_i^+ \bar{h}_j^+ - \bar{v}_i^- \bar{h}_j^- \right).
\end{equation}
According to \refeq{eq:point_symm}, any symmetric temporal window that is much shorter than $T$  can be used. 
For simplicity, we choose an exponential temporal window $K(\Delta)=\exp(\Delta/\tau_{STDP})$ with decay rate $\tau_{STDP} \ll T$ (\reffig{fig:CD_comparison}b). In this case, $\eta = 2A \frac{T-\tau_{br}}{2T} \tau_{STDP}$.

The modulatory function $g(t)$ partitions the training into several epochs of duration $2T$. Each epoch consists of a \ac{LTP} phase during which the data is presented (construction), followed by a free-running \ac{LTD} phase (reconstruction).
The weights are updated asynchronously during the time interval in which the neural sampling proceeds, and \refeq{eq:ecd_update} tells us that its average resembles \refeq{eq:cd_rule}. However, it is different in two ways: the averages are taken over one data and reconstruction phase rather than a mini-batch of data samples and their reconstructions; and more importantly, the synaptic weights are updated during the data and the reconstruction phase, whereas in the \ac{CD} rule, updates are carried out at the end of the reconstruction phase. In the derivation above the effect of the weight modification on the network during an epoch $2T$ was neglected for the sake of mathematical tractability. In the following, we verify that despite this approximation, the event-driven \ac{CD} performs nearly as well as standard \ac{CD} in a commonly used benchmark task.
\begin{figure}
\begin{center}
\includegraphics[width=0.48\textwidth]{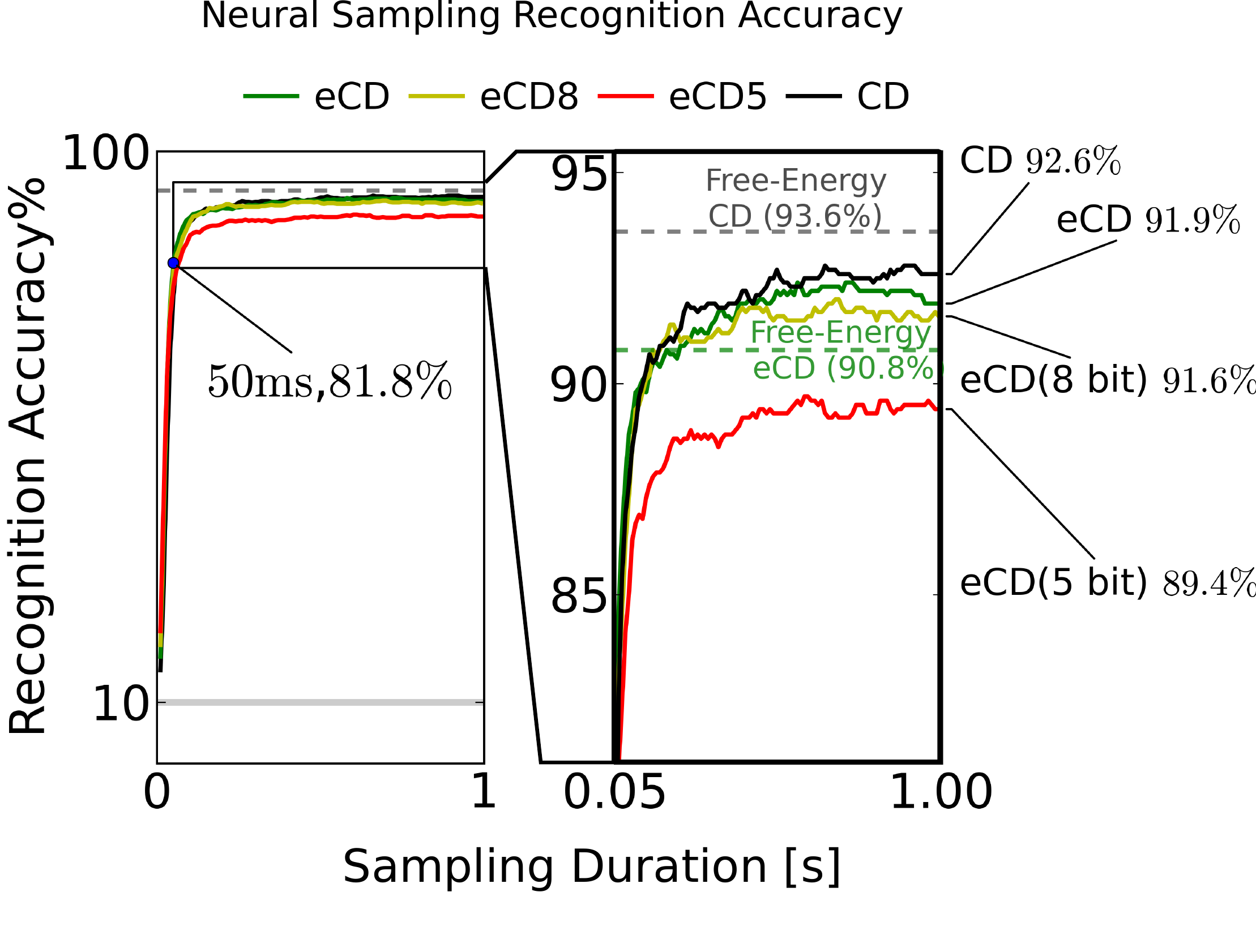}
  \begin{tabular}{l  c  c }        
                              & Accuracy & Accuracy\\
                              & Neural Sampler & Free-energy\\
    \hline
    Standard \ac{CD}                       & 92.6\%                      & 93.6\%                      \\ 
    Event-driven \ac{CD}                   & 91.9\%                      & 90.8\%                      \\ 
    Event-driven \ac{CD} (8 bits)          & 91.6\%                      & 91.0\%                      \\ 
    Event-driven \ac{CD} (5 bits)          & 89.4\%                      & 89.2\%                      \\ 
    \hline
  \end{tabular}
\end{center}
\caption{\label{tab:performance} To test recognition accuracy, the \acp{RBM} are sampled using the \ac{IF} neuron-based sampler for up to $\unit[1]{s}$. The classification is read out by identifying the group of class label neurons that had the highest activity. This experiment is run for \ac{RBM} parameter sets obtained by standard \ac{CD} (black, CD) and event-driven \ac{CD} (green, eCD). To test the robustness of the \ac{RBM}, it was run with parameters obtained by event-driven \ac{CD} discretized to 8 and 5 bits. In all scenarios, the accuracy after $\unit[50]{ms}$ of sampling was above $80\%$ and after $\unit[1]{s}$ the accuracies typically reached their peak at around $91.9\%$. The dashed horizontal lines show the recognition accuracy obtained by minimizing the free-energy (see text). The fact that the eCD curve (solid green) surpasses its free-energy minimization performance suggests that the \ac{RBM} learns a model that is tailored to the \ac{IF} spiking neural network.}
\end{figure}

\subsection{Learning a generative model of hand-written digits}
We train the \ac{RBM} to learn a generative model of the MNIST handwritten digits using event-driven \ac{CD} (see \refsec{sec:nnarch} for details). 
For training, $20000$ digits selected randomly from a training set consisting of 10000 digits were presented in sequence, with an equal number of samples for each digit. 

\begin{figure*}
   \begin{center}
   \includegraphics[width=1.0\textwidth]{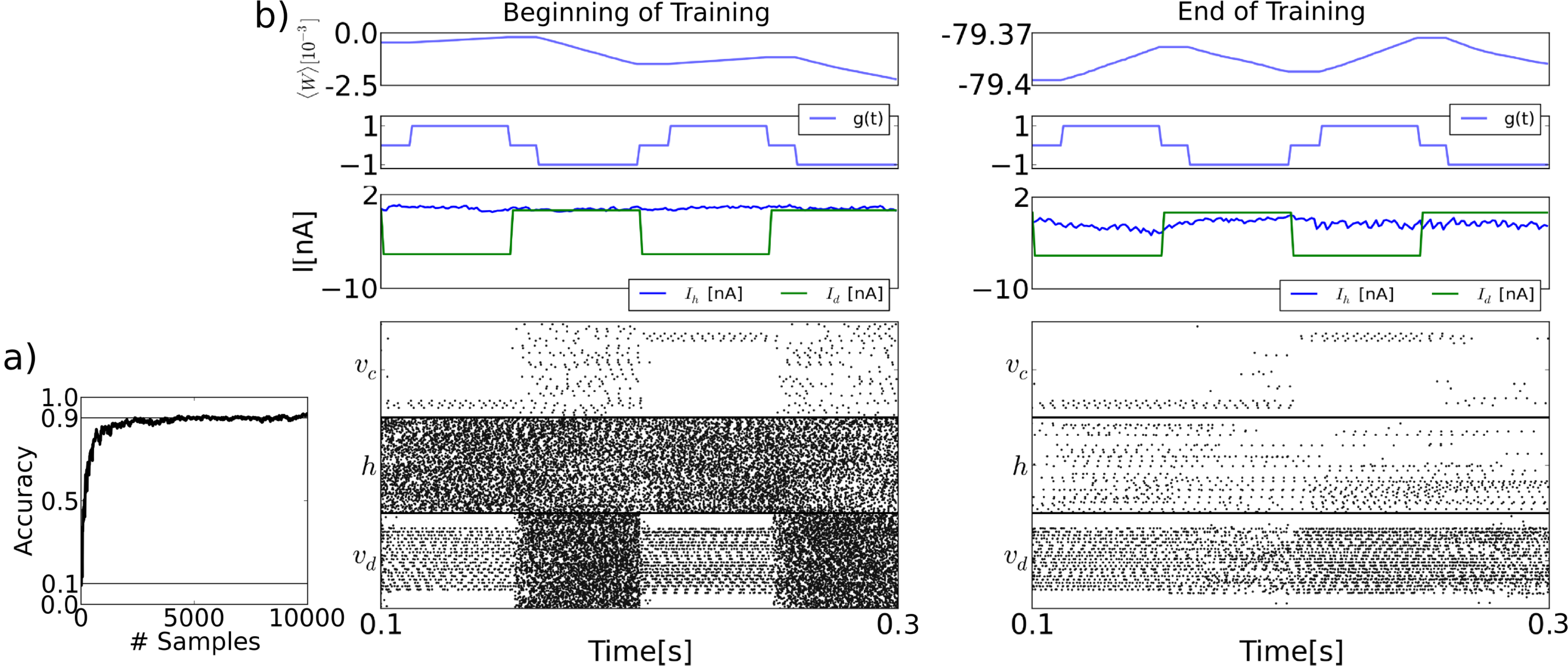}
   \caption{\label{fig:trained_rbm} The spiking neural network learns a generative model of the MNIST dataset using the event-driven \ac{CD} procedure. (a) Learning curve, shown here up to 10000 samples. (b) Details of the training procedure, before and after training (20000 samples). During the first half of each $\unit[.1]{s}$ epoch, the visible layer $v$ is driven by the sensory layer. During this phase, the gating variable $g$ is $1$, meaning that the synapses undergo \ac{LTP}. During the second half of each epoch, the sensory stimulus is removed, and $g$ is set to $-1$, so the synapses undergo \ac{LTD}. The top panels of both figures show the mean of the entries of the weight matrix. The second panel shows the values of the modulatory signal $g(t)$. The third panel shows the synaptic currents of a visible neuron, where $I_h$ is caused by the feedback from the hidden and the bias, and $I_d$ is the data. The timing of the clamping ($I_d$) and $g$ differ due to an interval $\tau_{br}$ where no weight update is undertaken to avoid the transients (See Materials and Methods). Before learning and during the reconstruction phase, the activity of the visible layer is random. But as learning progresses, the activity in the visible layer reflects the presented data in the reconstruction phase. This is very well visible in the layer class label neurons $v_c$, whose activity persists after the sensory stimulus is removed.
   Although the firing rates of the hidden layer neurons before training is high (average $\unit[113]{Hz}$), this is only a reflection of the initial conditions for the recurrent couplings $W$. In fact, at the end of the training, the firing rates in both layers becomes much sparser (average $\unit[9.31]{Hz}$).}
   \end{center}
\end{figure*}

The raster plots in \reffig{fig:trained_rbm} show the spiking activity of each layer before and after learning for epochs of duration $\unit[100]{ms}$.
The top panel shows the population-averaged weight.
After training, the sum of the upwards and downward excursions of the average weight is much smaller than before training, because the learning is near convergence.
The second panel shows the value of the modulatory signal $g(t)$.
The third panel shows the input current ($I_d$) and the current caused by the recurrent couplings ($I_h$).

Two methods to estimate the overall classification performance of the neural \ac{RBM} can be used. The first is by neural sampling: the visible layer is clamped to the digit only, and the network is run for $1\unit{s}$. The known label is then compared with the positions of the group of class neurons that had the highest population rate.
The second method is by minimizing free-energy: the neural \acp{RBM} parameters are extracted, and for each data sample, the class neurons with the lowest free-energy (See Materials and Methods) is compared to the  known label.
In both cases, recognition was tested for $1000$ data samples that were not used during the training.
The results are summarized in \reffig{tab:performance}.

As a reference we provide the best performance achieved using the standard \ac{CD} and one unit per class label ($N_c=10$) (\reffig{tab:performance}, table row 1), $93.6\%$. By mapping the learned parameters to the neural \ac{RBM} the recognition accuracy reached $92.6\%$.

When training a neural \ac{RBM} of \ac{IF} neurons using event-driven \ac{CD}, the recognition result was $91.9\%$ (\reffig{tab:performance}, table row 2).
The performance of this \ac{RBM} obtained by minimizing its free-energy was $90.8\%$.
The learned parameters performed well for classification using the free-energy calculation which suggests that the network learned a model that is consistent with the mathematical description of the \ac{RBM}.

In an energy-based model like the \ac{RBM} the free-energy minimization should give the upper bound on the discrimination performance \citepd{Haykin99}. For this reason, the fact that the recognition accuracy is higher when sampling as opposed to using the free-energy method may appear puzzling.
However, this is possible because the neural \ac{RBM} does not exactly sample from the Boltzmann distribution, as explained in \refsec{sec:kldivergence}. This suggests that event-driven \ac{CD} compensates for the discrepancy between the distribution sampled by the neural \ac{RBM} and the Boltzmann distribution, by learning a model that is tailored to the spiking neural network.

Excessively long training durations can be impractical for real-time neuromorphic systems.
Fortunately, the learning using event-driven \ac{CD} is fast: 
Compared to the off-line \ac{RBM} training ($250000$ presentations, in mini-batches of $100$ samples) the event-driven \ac{CD} training succeeded with a smaller number of data presentations ($20000$), which corresponded to $\unit[2000]{s}$ of simulated time.
This suggests that the training durations are achievable for real-time neuromorphic systems.

\paragraph{The choice of the number of class neurons $N_c$.} Event-driven \ac{CD} underperformed in the case of $1$ neuron per class label ($N_c=10$), which is the common choice for standard \ac{CD} and Gibbs sampling. 
This is because a single neuron firing at its maximum rate of $\unit[250]{Hz}$ cannot efficiently drive the rest of the network without tending to induce spike-to-spike correlations (\emph{e.g.} synchrony), which is incompatible with the assumptions made for sampling with \ac{IF} neurons and event-driven \ac{CD}. As a consequence, the generative properties of the neural \ac{RBM} degrade.
This problem is avoided by using several neurons per class label (in our case four neurons per class label)  because the synaptic weight can be much lower to achieve the same effect, resulting in smaller spike-to-spike correlations.

\subsection{Generative properties of the \ac{RBM}}
We test the neural \ac{RBM} as a generative model of the MNIST dataset of handwritten digits, using parameters obtained by running the event-driven \ac{CD}.

\begin{figure*}
   \begin{center}
   \includegraphics[width=.8\textwidth]{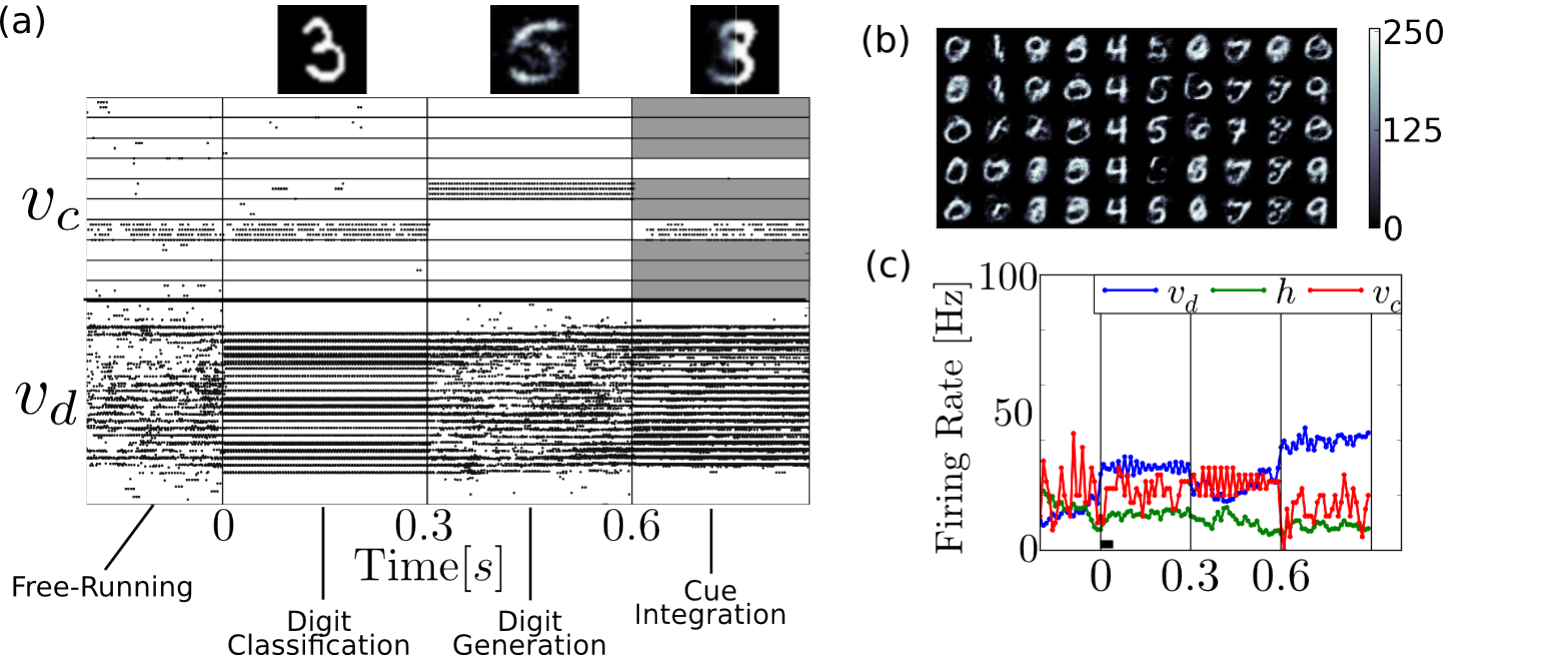} 
   \caption{\label{fig:classification_reconstruction_inference} The recurrent structure of the network allows it to classify, reconstruct and infer from partial evidence. (a) Raster plot of an experiment illustrating these features. Before time $0s$, the neural \ac{RBM} runs freely, with no input. Due to the stochasticity in the network, the activity wanders from attractor to attractor. At time $0s$, the digit $3$ is presented (\emph{i.e.} layer $v_d$ is driven by $d$), activating the correct class label in $v_c$; At time $t=\unit[.3]{s}$, the class neurons associated to $5$ are clamped to high activity and the rest of the class label neurons are strongly inhibited, driving the network to reconstruct its version of the digit in layer $v_d$; At time $t=\unit[.6]{s}$, the right-half part of a digit 8 is presented, and the class neurons are stimulated such that only $3$ or $6$ can activate (all others are strongly inhibited as indicated by the gray shading). Because the stimulus is inconsistent with $6$, the network settles to a $3$ and attempts to reconstruct it. The top figures shows the digits reconstructed in layer $v_d$. (b) Digits $0-9$, reconstructed in the same manner. Each row corresponds to a different, independent run. (c) Population firing rate of the experiment presented in (a). During recognition, the network typically reaches equilibrium after about $10\tau_r=\unit[40]{ms}$ (black bar).}
   \end{center}
\end{figure*}

In the context of the handwritten digit task, the \ac{RBM}'s generative property enables it to classify digits, generate them, and infer a digit by combining partial evidence.
These features are clearly illustrated in the following experiment (\reffig{fig:classification_reconstruction_inference}). First the digit $3$ is presented (\emph{i.e.} layer $v_d$ is driven by layer $d$) and the correct class label in $v_c$ activated. Second, the neurons associated to class label $5$ are clamped, and the network generated its learned version of the digit. Third, the right-half part of a digit $8$ is presented, and the class neurons are stimulated such that only $3$ or $6$ are able to activate (the other class neurons are inhibited, indicated by the gray shading). Because the stimulus is inconsistent with $6$, the network settled to $3$ and reconstructed the left part of the digit.

The latter part of the experiment illustrates the integration of information between several partially specified cues, which is of interest for solving sensorimotor transformation or multi-modal sensory cue integration problems \citepd{Deneve_etal01,Doya_etal07,Corneil_etal12b}. This feature has been used for auditory-visual sensory fusion in a spiking \ac{DBN} model \citepd{OConnor_etal13}. There, the authors trained a \ac{DBN} with visual and auditory data, which learned to associate the two sensory modalities, very similarly to how class labels and visual data are associated in our architecture. Their network was able to resolve a similar ambiguity as in our experiment in \reffig{fig:classification_reconstruction_inference}, but using auditory inputs instead of a class label.

In the digit generation mode, the trained network had a tendency to be globally bistable, whereby the layer $v_d$ completely deactivated layer $h$. Since all the interactions between $v_d$ and $v_c$ take place through the hidden layer, $v_c$ could not reconstruct the digit. 
To avoid this, we added two populations of \ac{IF} neurons that were wired to layers $v$ and $h$, respectively. The parameters of these neurons and their couplings were tuned such that each layer was strongly excited when it's average firing rate fell below $\unit[5]{Hz}$.

\paragraph{Neural parameters with finite precision.}
In hardware systems, the parameters related to the weights and biases cannot be set with floating-point precision, as can be done in a digital computer. In current neuromorphic implementations the synaptic weights can be configured at precisions of about $8$ bits \citepd{Yu_Cauwenberghs10b}. 
We characterize the impact of finite-precision synaptic weights on performance by discretizing the weight and bias parameters to $8$ bits and $5$ bits. The set of possible weights were spaced uniformly in the interval $(\mu-4.5 \sigma, \mu+4.5 \sigma)$, where $\mu,\sigma$ are the mean and the standard deviation of the parameters across the network, respectively.
The classification performance of MNIST digits degraded gracefully. In the $8$ bit case, it degrades only slightly to $91.6\%$, but in the case of $5$ bits, it degrades more substantially to $89.4\%$. In both cases, the \ac{RBM} still retains its discriminative power, which is encouraging for implementation in hardware neuromorphic systems.
\section{Discussion}
Neuromorphic systems are promising alternatives for large-scale implementations of \acp{RBM} and deep networks, but the common procedure used to train such networks, \acf{CD}, involves iterative, discrete-time updates that do not straightforwardly map on a neural substrate.
We solve this problem in the context of the \ac{RBM} with a spiking neural network model that uses the recurrent network dynamics to compute these updates in a continuous-time fashion. We argue that the recurrent activity coupled with \ac{STDP} dynamics implements an event-driven variant of \ac{CD}.
Using event-driven \ac{CD}, the network connectivity remains unchanged during training and testing, enabling the system to learn in an on-line fashion, while being able to carry out functionally relevant tasks such as recognition, data generation and cue integration.

The \ac{CD} algorithm can be used to learn the parameters of probability distributions other than the Boltzmann distribution (even those without any symmetry assumptions). Our choice for the \ac{RBM}, whose underlying probability distribution is a special case of the Boltzmann distribution, is motivated by the following facts: They are universal approximators of discrete distributions \citepd{Le-Roux_Bengio08}; the conditions under which a spiking neural circuit can naturally perform MCMC sampling of a Boltzmann distribution were previously studied \citepd{Merolla_etal10,Buesing_etal11}; and \acp{RBM} form the building blocks of many deep learning models such as \acp{DBN}, which achieve state-of-the-art performance in many machine learning tasks \citepd{Bengio09}. The ability to implement \acp{RBM} with spiking neurons and train then using event-based \ac{CD} paves the way towards on-line training of \acp{DBN} of spiking neurons \citepd{Hinton_etal06}.

We chose the MNIST handwritten digit task as a benchmark for testing our model. When the \ac{RBM} was trained with standard \ac{CD}, it could recognize up to 926 out of 1000 of out-of-training samples. 
The MNIST handwritten digits recognition task was previously shown in a digital neuromorphic chip \citepd{Arthur_etal12}, which performed at $89\%$ accuracy, and in a software simulated visual cortex model \citepd{Eliasmith_etal12}. However, both implementations were configured using weights trained off-line.
A recent article showed the mapping of off-line trained \acp{DBN} onto spiking neural network \citepd{OConnor_etal13}. Their results demonstrated hand-written digit recognition using neuromorphic event-based sensors as a source of input spikes. Their performance reached up to $94.1\%$ using leaky \ac{IF} neurons. The use of off-line \ac{CD} combined with an additional layer explains to a large extent their better performance compared to ours. Our work extends \citepd{OConnor_etal13} by demonstrating an on-line training using synaptic plasticity, testing its robustness to finite weight precision, and providing an interpretation of spiking activity in terms of neural sampling.

To achieve the computations necessary for sampling from the \ac{RBM}, we have used the neural sampling framework \citepd{Fiser_etal10}, where each spike is interpreted as a sample of an underlying probability distribution. \citeauthor{Buesing_etal11} proved that abstract neuron models consistent with the behavior of biological spiking neurons can perform \ac{MCMC}, and have applied it to a basic learning task in a fully visible Boltzmann Machine.
We extended the neural sampling framework in three ways: First, we identified the conditions under which a dynamical system consisting of \ac{IF} neurons can perform neural sampling; Second, we verified that the sampling of \acp{RBM} was robust to finite-precision parameters; Third, we demonstrated learning in a Boltzmann Machine with hidden units using \ac{STDP} synapses.

In the neural sampling framework, neurons behave stochastically. This behavior can be achieved in \ac{IF} neurons using noisy input currents, created by a Poisson spike train. 
Spike trains with Poisson-like statistics can be generated with no additional source of noise, for example by the following mechanisms: balanced excitatory and inhibitory connections \citepd{Vreeswijk_Sompolinsky96}, finite-size effects in a large network, and neural mismatch \citepd{Amit_Brunel97}.
The latter mechanism is particularly appealing, because it benefits from fabrication mismatch and operating noise inherent to neuromorphic implementations~\citepd{Chicca_Fusi01}. 

Other groups have also proposed to use \ac{IF} neuron models for computing the Boltzmann distribution.
\citepd{Merolla_etal10} have shown that noisy \ac{IF} neurons' activation function is approximately sigmoidal as required by the Boltzmann machine, and have devised a scheme whereby a global inhibitory rhythm drives the network to generate samples of the Boltzmann distribution. 
\citepd{OConnor_etal13} have demonstrated a deep belief network of \ac{IF} neurons that was trained off-line, using standard \ac{CD} and tested it using the MNIST database.
Independently and simultaneously to this work, \citepd{Petrovici_etal13} demonstrated that conductance-based \ac{IF} neurons in a noisy environment are compatible with neural sampling as described in \citepd{Buesing_etal11}. Similarly, \citeauthor{Petrovici_etal13} find that the choice of non-rectangular \acp{PSP} and the approximations made by the \ac{IF} neurons are not critical to the performance of the neural sampler. 
Our work extends all of those above by providing an online, \ac{STDP}-based learning rule to train \acp{RBM} sampled using \ac{IF} neurons.

\paragraph{Applicability to neuromorphic hardware.} 
Neuromorphic systems are sensible to fabrication mismatch and operating noise. 
Fortunately, the mismatch in the synaptic weights and the activation function parameters $\gamma$ and $\beta$ are not an issue if the biases and the weights are learned, and the functionality of the \ac{RBM} is robust to small variations in the weights caused by discretization.
These two findings are encouraging for neuromorphic implementations of \acp{RBM}.
However, at least two conceptual problems of the presented \ac{RBM} architecture must be solved in order to implement such systems on a large-scale.
First, the symmetry condition required by the \ac{RBM} does not necessarily hold.  
In a neuromorphic device, the symmetry condition is impossible to guarantee if the synapse weights are stored locally at each neuron.
Sharing one synapse circuit per pair of neurons can solve this problem.
This may be impractical due to the very large number of synapse circuits in the network, but may be less problematic when using \acp{RRAM} (also called \emph{memristors}) crossbar arrays to emulate synapses \citepd{Cruz-Albrecht_etal13,Kuzum_etal11,Serrano-Gotarredona_etal13}.\ac{RRAM} are a new class of nanoscale devices whose current-voltage relationship depends on the history of other electrical quantities \citepd{Strukov_etal08}, and so act like programmable resistors. Because they can conduct currents in both directions, one \ac{RRAM} circuit can be shared between a pair of neurons. 
A second problem is the number of recurrent connections. Even our \ac{RBM} of modest dimensions involved almost 2 million synapses, which is impractical in terms of bandwidth and weight storage. Even if a very high number of weights are zero, the connections between each pair of neurons must exist in order for a synapse to learn such weights. 
One possible solution is to impose sparse connectivity between the layers \citepd{Tang_Eliasmith10,Murray_Kreutz-Delgado07}. This remains to be tested in our model.
\paragraph{Outlook: A custom learning rule.} Our method combines \ac{IF} neurons that perform neural sampling and the \ac{CD} rule.
Although we showed that this leads to a functional model, we do not know whether event-driven \ac{CD} is optimal in any sense. 
This is partly due to the fact that \ac{CD}$_k$ is an approximate rule \citepd{Hinton02}, and it is still not entirely understood why it performs so well, despite extensive work in studying its convergence properties \citepd{Carreira-Perpinan_Hinton05}.
Furthermore, the distribution sampled by the \ac{IF} neuron does not exactly correspond to the Boltzmann distribution, and the average weight updates in event-driven \ac{CD} differ from those of standard \ac{CD}, because in the latter they are carried out at the end of the reconstruction step.

A very attractive alternative is to derive a custom synaptic plasticity rule that minimizes some functionally relevant quantity (such as Kullback-Leibler divergence or Contrastive Divergence), \emph{given} the encoding of the information in the \ac{IF} neuron \citepd{Deneve08,Brea_etal13}. 
A similar idea was recently pursued in \citepd{Brea_etal13}, where the authors derived a triplet-based synaptic learning rule that minimizes an upper bound of the Kullback-Leibler divergence between the model and the data distributions. 
Interestingly, their rule had a similar global signal that modulates the learning rule, as in event-driven \ac{CD}, although the nature of this resemblance remains to be explored. 
Such custom learning rules can be very beneficial in guiding the design of on-chip plasticity in neuromorphic \ac{VLSI} and \ac{RRAM} nanotechnologies, and will be the focus of future research.

\section*{Acknowledgments}

This work was partially funded by the National Science Foundation (NSF EFRI-1137279), the Office of Naval Research (ONR MURI 14-13-1-0205), and the Swiss National Science Foundation (PA00P2\_142058). We thank all the anonymous reviewers of a previous version of this article for their constructive comments.




\end{document}